\documentclass[journal,twoside,web]{ieeecolor}
\usepackage{generic}
\usepackage{cite}
\usepackage{graphicx}
\usepackage{hyperref}
\usepackage{textcomp}
\usepackage{amsmath,amssymb,amsfonts}
\usepackage{algorithmic,float}
\usepackage[ruled]{algorithm2e}
\SetKwInput{KwInput}{Input}                % Set the Input
\SetKwInput{KwOutput}{Output}              % set the Output
\SetKwInput{KwRepeat}{Repeat}
\SetKwInput{KwUntil}{Until}
\SetKwInput{KwReturn}{Return}
\usepackage{amsfonts}
\usepackage{graphicx}
\usepackage{textcomp}
\usepackage{tabularx}
\usepackage{array}
\usepackage{booktabs}
\usepackage{threeparttable}  
\usepackage{multirow}
\usepackage{bm}
\usepackage{amsmath}
\usepackage{subfigure}
\usepackage{amssymb}% http://ctan.org/pkg/amssymb
\usepackage{pifont}% http://ctan.org/pkg/pifont
\usepackage{color}

\newcommand{\Romann}[1]{\uppercase\expandafter{\romannumeral#1}}
\newcommand{\codezs}[1]{\textcolor[RGB]{0,129,0}{\tcc{#1}}}

\def\BibTeX{{\rm B\kern-.05em{\sc i\kern-.025em b}\kern-.08em
    T\kern-.1667em\lower.7ex\hbox{E}\kern-.125emX}}
\begin{document}
\title{BiBLDR: Bidirectional Behavior Learning for Drug Repositioning}
\author{Renye Zhang, Mengyun Yang, Qichang Zhao, Jianxin Wang
\thanks{Mengyun Yang is the Corresponding Author. E-mail: mengyun\_yang@126.com}
\thanks{Renye Zhang and Mengyun Yang are with the School of Computer Science, Hunan First Normal University, Hunan, China, 410205. Qichang Zhao and Jianxin Wang are with the School of Computer Science and Engineering, Central South University, Hunan, China, 410083.}
\thanks{This work was supported by the National Natural Science Foundation of China (Grant no. 62272309).}
}
\maketitle
\begin{abstract}
Drug repositioning aims to identify potential new indications for existing drugs to reduce the time and financial costs associated with developing new drugs. Most existing deep learning-based drug repositioning methods predominantly utilize graph-based representations. However, graph-based drug repositioning methods struggle to perform effective inference in cold-start scenarios involving novel drugs because of the lack of association information with the diseases. Unlike traditional graph-based approaches, we propose a bidirectional behavior learning strategy for drug repositioning, known as BiBLDR. This innovative framework redefines drug repositioning as a behavior sequential learning task to capture drug-disease interaction patterns. First, we construct bidirectional behavioral sequences based on drug and disease sides. The consideration of bidirectional information ensures a more meticulous and rigorous characterization of the behavioral sequences. Subsequently, we propose a two-stage strategy for drug repositioning. In the first stage, we construct prototype spaces to characterize the representational attributes of drugs and diseases. In the second stage, these refined prototypes and bidirectional behavior sequence data are leveraged to predict potential drug-disease associations. Based on this learning approach, the model can more robustly and precisely capture the interactive relationships between drug and disease features from bidirectional behavioral sequences. Extensive experiments demonstrate that our method achieves state-of-the-art performance on benchmark datasets. Meanwhile, BiBLDR demonstrates significantly superior performance compared to previous methods in cold-start scenarios. Our code is published in https://github.com/Renyeeah/BiBLDR.
\end{abstract}


\section{Introduction}
\label{sec:introduction}

\IEEEPARstart{D}{rug} repositioning seeks to identify new therapeutic indications for existing, approved drugs, thereby uncovering potential treatments for diseases that may respond to these agents~\cite{huang2024foundation}. The traditional cold-start drug development process is generally divided into three phases, each demanding significant expenditures and prolonged research timelines. Compared to the conventional drug development process, drug repositioning circumvents the need for Phase I clinical trials—which focus on safety evaluation—by utilizing existing clinical safety data from approved drugs. This approach substantially reduces the financial costs and temporal demands of bringing a therapeutic agent to market~\cite{farha2019drug}. The pharmaceutical industry has witnessed numerous clinically and commercially significant successes in drug repositioning, including aspirin~\cite{ning2024research} (repurposed from an analgesic to a cardiovascular prophylactic), minoxidil~\cite{zhang2024culprits} (redeveloped from an antihypertensive to treatment for alopecia), and remdesivir~\cite{gottlieb2024oral} (transitioned from an investigational antiviral to therapy for COVID-19).

Computational drug repositioning involves the systematic identification of potential indications for drugs through the development of mathematical models and the design of algorithms, rather than relying solely on incidental discoveries made during clinical practice. SCMFDD~\cite{zhang2018predicting} projects the drug-disease association relationship into two low-rank spaces,and then introduces drug feature-based similarity and disease semantic similarity as constraints. iDrug~\cite{chen2020idrug} incorporates graph regularization into the decomposition of the drug–disease association matrix to enhance the interpretability of the results. BNNR~\cite{yang2019drug} integrates association and similarity matrices and performs matrix completion under the low-rank assumption, which can naturally handle the cold-start problem. NRLMF~\cite{liu2016neighborhood} introduces local regularization on the local structure of drug-disease associations to obtain semantically richer matrix decomposition results. With the support of large-scale biomedical data and advancements in computational hardware, deep learning-based computational drug repositioning has garnered significant attention in recent years. deepDR~\cite{zeng2019deepdr} learns shallow drug representations from heterogeneous networks using a multimodal deep autoencoder, followed by extracting high-level features through a VAE to predict drug-disease associations. DDAGDL~\cite{zhao2022geometric} combines the biological attributes of drugs and diseases with geometric network priors, using geometric deep learning to project non-Euclidean biomedical data into a latent feature space, capturing drug and disease features more comprehensively. 

Drug-disease associations and similarity data represent graph topological structures, so many studies conceptualize the drug repositioning task as a link prediction problem within graphs. This contributes to the widespread use of \emph{Graph Convolutional Networks} (GCN) in deep learning-based methods. NIMCGCN~\cite{li2020neural} uses GCN to extract deep features of entities and then uses neural induction matrix filling to predict association values. DRWBNCF~\cite{meng2022weighted} uses a weighted bilinear graph convolution operation to integrate association and similarity networks into a unified representation, and uses multilayer perceptrons and graph regularization to model drug-disease associations. PSGCN~\cite{sun2022partner} generates independent subgraphs for each drug-disease pair to be predicted, transforming the link prediction task into a graph classification problem. This approach enhances the model's sensitivity to local topological information. AdaDR~\cite{sun2024drug} overcomes the challenge of insufficient integration between features and topological structures in traditional GCNs for drug repositioning tasks. It simultaneously extracts embeddings from both the node feature and topological space, employing adaptive graph convolution to model their interactions. 

However, in addition to repositioning existing drugs, repurposing newly developed drugs remains a major challenge. These compounds possess very limited a priori knowledge and lack established associations with diseases. When represented within graph structures, the corresponding nodes often exhibit sparse connectivity, with few associated nodes and edges. Consequently, in cold-start scenarios of drug repositioning involving novel drugs, GCN-based approaches often struggle to perform effective inference due to the absence of association information.

\begin{figure}
    \centering
    \includegraphics[width=1.0\linewidth]{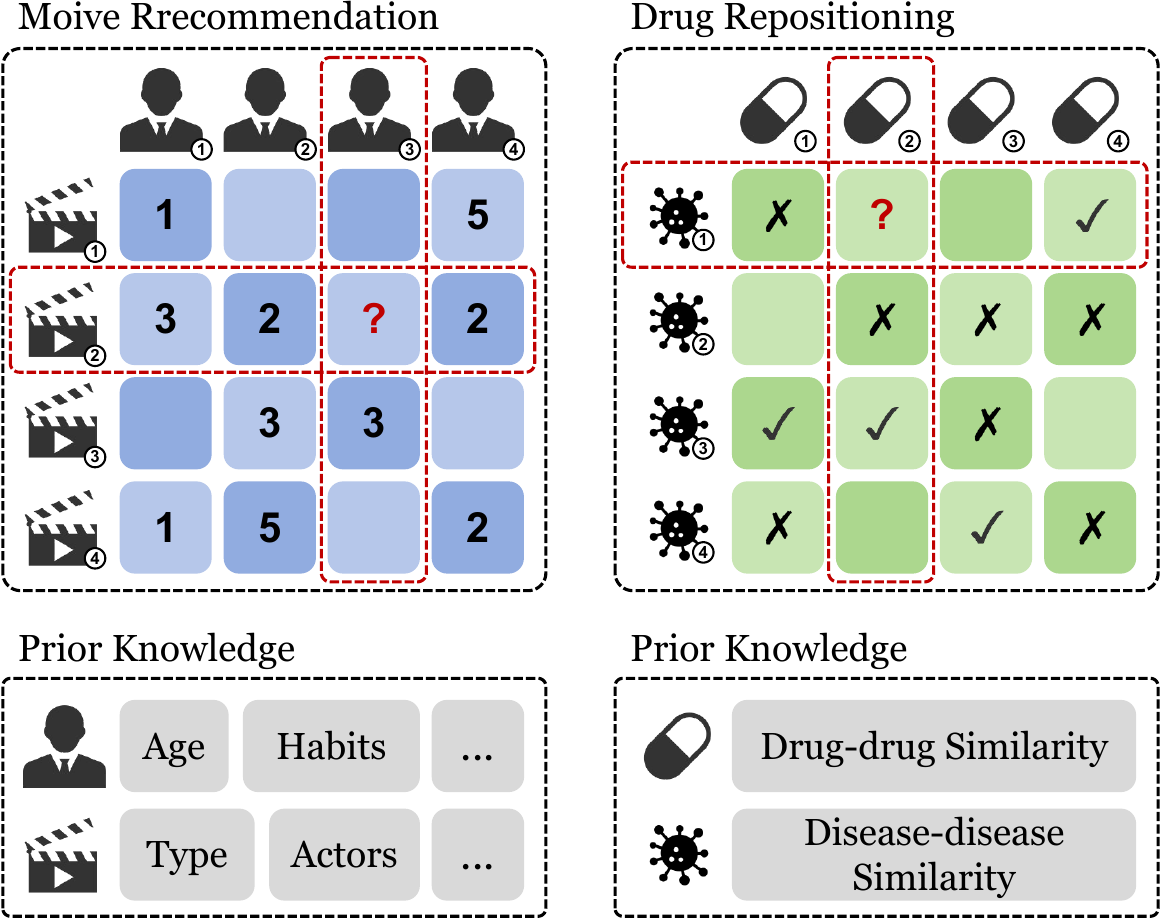}
    \caption{The comparative task frameworks on normal movie recommendation and drug repositioning. The blue box represents the user's movie rating from 1 to 5. The green box indicates the drug's therapeutic effect on the disease.}
    \label{Introduction}
\end{figure}

Given the evident behavioral relationships between drugs and diseases, the drug repositioning task can be defined as a recommendation system~\cite{wang2015mining, ozsoy2018realizing}. The recommendation system is a technique that predicts content a user may find interesting by analyzing their habits and historical behavior~\cite{zhang2019deep}. It is widely used in fields such as e-commerce platforms~\cite{shankar2024intelligent} and social media~\cite{sharma2024survey}, assisting users in discovering potentially relevant content. For instance, in the context of movie recommendation systems, user preferences are inferred from historical viewing records, termed a behavioral sequence, to generate personalized recommendations for new films. Analogously, in computational drug repositioning, a drug’s approved therapeutic indications can be conceptualized as its behavioral sequence, which is subsequently leveraged to infer novel drug-disease associations through predictive modeling. Fig.~\ref{Introduction} provides a comparative schematic of the methodological frameworks applied to both tasks.

As shown in Fig.~\ref{Introduction}, the objective of the movie recommendation task is to use the user-movie rating matrix to predict the user's rating for a particular movie, thereby determining whether the system should recommend that movie to the user. Similarly, the drug repositioning task utilizes the drug-disease association matrix to predict the potential therapeutic effect of a drug on a specific disease. The prior knowledge used in the former includes the attributes of users and movies, while in this paper, the prior knowledge used in the latter consists of similar data between drugs and diseases. Both the approved therapeutic indications of a drug and the known sensitivity of a disease to the drug can be treated as behavioral sequence data. The detailed process of behavioral sequence modeling is discussed in Section~\ref{2.2}. Deep learning techniques make significant advancements in the field of recommendation systems~\cite{guo2017deepfm, xie2022contrastive}. In recent years, the Transformer architecture~\cite{vaswani2017attention} has achieved remarkable success in natural language processing~\cite{kenton2019bert, brown2020language, achiam2023gpt}. The multi-head attention mechanism employed by the Transformer offers significant advantages in handling sequential data. Consequently, many studies apply the Transformer to recommendation systems, enabling it to process user behavior sequence data effectively. BERT4Rec~\cite{sun2019bert4rec} enhances the representational power of behavior sequence by introducing a deep bilateral self-attention mechanism, allowing each position to integrate bidirectional information. BST~\cite{chen2019behavior} captures temporal dependencies in user behavior sequences, overcoming traditional methods' reliance on feature concatenation or unidirectional attention, enabling more accurate modeling of item interactions. PBAT~\cite{su2023personalized} introduces a novel behavior-aware attention mechanism by integrating behavioral relevance and personalized patterns. This mechanism simultaneously considers the impact of behavior, time, and location factors on sequence collaboration. 

Inspired by the aforementioned works, to address the limitations of traditional graph representations in drug repositioning tasks, we propose \emph{\textbf{\underline{Bi}}directional \textbf{\underline{B}}ehavior \textbf{\underline{L}}earning for \textbf{\underline{D}}rug \textbf{\underline{R}}epositioning} (BiBLDR), which analyzes drug-disease association data from the perspective of behavioral sequences. In addition, it is the first to apply a Transformer-based architecture to process behavioral sequence data in drug repositioning tasks. Furthermore, our approach is based on a two-stage training process. The first stage uses similarity data to construct prototype spaces for drugs and diseases independently, while the second stage predicts the association between drugs and diseases by analyzing their bidirectional behavioral sequences. The contributions of this paper are as follows:

\begin{itemize}
    \item We propose a novel learning paradigm for drug repositioning: bidirectional behavioral sequence analysis. This strategy simultaneously constructs behavioral sequence data from both drug and disease aspect, integrating drugs and diseases into a unified recommendation system framework, leveraging bidirectional information allows the model to effectively mitigate the cold-start problem by preserving a part of behavioral signals.
    \item We propose a two-stage drug repositioning strategy. In the first stage, we employ similarity-driven learning to establish prototype spaces for drugs and diseases. This spatial organization translates entity similarities into geometrically constrained vector relationships, enabling the model to extract structural relationships within the latent prototype space. The subsequent stage integrates these prototypical representations with bidirectional behavioral sequences to predict drug-disease associations, with bidirectional information augmenting feature expressiveness through multi-domain representation fusion.
    \item We are the first to apply the Transformer architecture to drug repositioning tasks based on behavioral sequence prediction, utilizing the multi-head attention mechanism to capture dependencies between drug and disease behavioral sequences, which synergistically strengthens the interaction of drug and disease features.
    \item We conduct extensive experiments on public datasets, and the results show that our BiBLDR significantly outperforms existing methods.
\end{itemize}

The organization of the remaining sections of this paper is as follows: Section~\ref{Section2} proposes the construction method of bidirectional behavioral sequences and provides a detailed framework of BiBLDR. Section~\ref{Section3} describes the datasets, experimental implementation details, and comprehensive results and analysis. Finally, the conclusion is presented in Section~\ref{Section4}.

\section{Methodology}\label{Section2}
This section commences with an overview of the methodological framework underpinning BiBLDR, followed by a systematic presentation of the proposed data modeling strategy for bidirectional behavioral sequences in computational drug repositioning. Subsequently, a comprehensive exposition of the framework is presented, detailing its architectural components, operational mechanisms, and theoretical underpinnings.

\subsection{Overall Framework}
\begin{figure*}[htbp]
    \centering
    \includegraphics[width=1.0\linewidth]{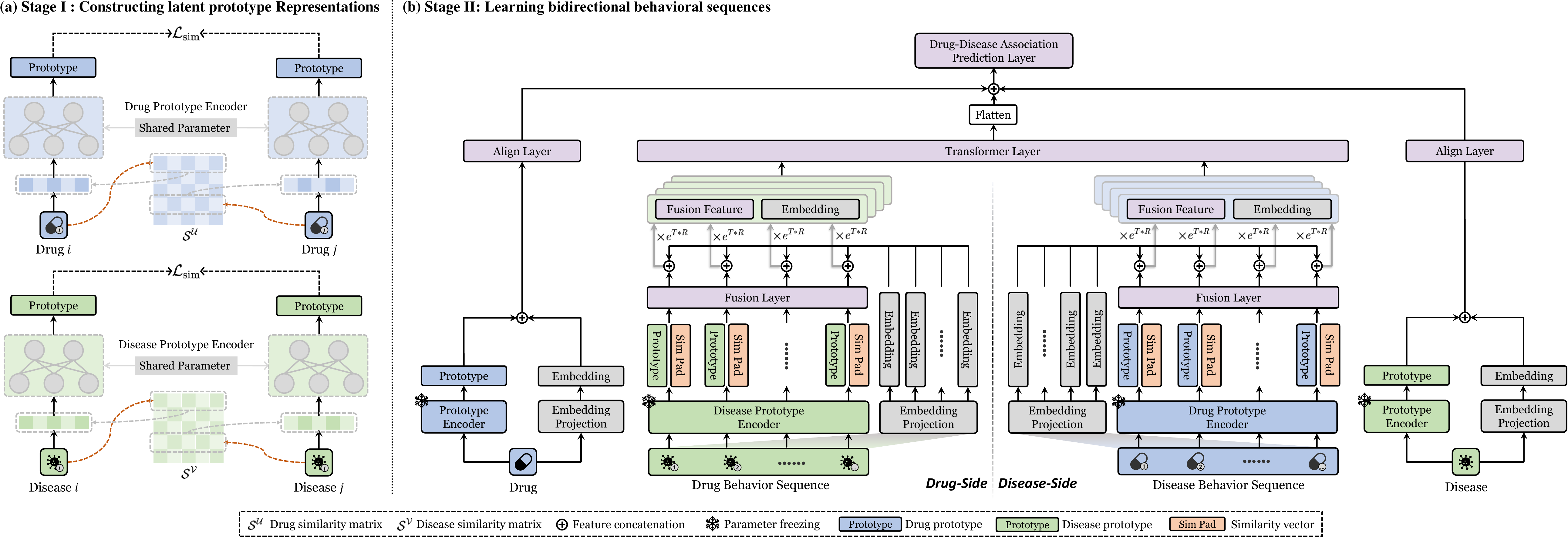}
    \caption{The proposed BiBLDR framework. (a) Utilize similarity data to construct prototype spaces for drugs and diseases separately. (b) Utilize prototypes and bidirectional behavioral sequence information to predict drug-disease associations.}
    \label{main}
\end{figure*}

Unlike conventional graph-based representation learning approaches, our methodology redefines drug repositioning as a behavioral sequence recommendation task. Our framework transmutes static matrix structures into behavioral sequences and proposes a two-stage learning strategy to perform drug repositioning task. In the first stage, drugs and diseases are represented as feature vectors derived from their similarity sequences. The Siamese neural network refines these representations by minimizing discrepancies between predicted and actual similarity scores, generating optimized prototype spaces for both entities. The second stage processes behavioral sequences of drugs and diseases in parallel, integrating similarity information to learn enriched semantic representations from bidirectional behavior sequences. Simultaneously, domain-specific attribute features are extracted through specialized pipelines. The entity-attribute and bidirectional behavioral representations are integrated to generate the final association prediction scores. Fig.~\ref{main} provides an overview of the overall framework.

\subsection{Bidirectional Behavioral Sequence Modeling}\label{2.2}
Traditional behavioral sequence prediction tasks focus on unidirectional user-item interactions. By contrast, our work jointly models the behavioral sequences of drugs and diseases to capture their bidirectional relationships. Specifically, our framework accounts for a drug’s therapeutic effects on diseases and a disease’s sensitivity to drugs. This bidirectional modeling enhances sequence representation by integrating drug-disease interdependencies. This section provides a detailed description of the construction method for the drug-disease bidirectional behavior sequence.

\subsubsection{Dataset Format} 
The dataset used contains a drug-disease association matrix $\mathcal{A} \in \mathbb{R}^{|\mathcal{U}| \times |\mathcal{V}|}$, a drug-drug similarity matrix $\mathcal{S}^\mathcal{U} \in \mathbb{R}^{|\mathcal{U}| \times |\mathcal{U}|}$, a disease-disease similarity matrix $\mathcal{S}^\mathcal{V} \in \mathbb{R}^{|\mathcal{V}| \times |\mathcal{V}|}$. $|\mathcal{U}|$ and $|\mathcal{V}|$ represent the total number of drugs and diseases. $\mathcal{A}_{ij} \in \{0, 1\}$, indicating the therapeutic relationship between $i$-th drug and $j$-th disease. All elements of $\mathcal{A}$ are divided into the training set and test set, with only the elements in the training set being accessible.  $\mathcal{S}_{ij}^{\mathcal{U}} \in [0, 1]$, indicating the similarity between $i$-th and $j$-th drug. $\mathcal{S}_{ij}^{\mathcal{V}} \in [0, 1]$, indicating the similarity between $i$-th and $j$-th disease.

\subsubsection{Reconceptualize Behavioral Sequence}
We define a scenario that uses behavioral sequences to model the drug repositioning task. Let $\mathcal{U} = \{u_1, u_2, \cdots, u_{|\mathcal{U}|}\}$ and $\mathcal{V} = \{v_1, v_2, \cdots, v_{|\mathcal{V}|} \}$. $\mathcal{U}$ and $\mathcal{V}$ represent the sets of all drugs and diseases in the dataset, respectively. $u_i$ and $v_j$ represent the $i$-th drug and $j$-th disease. Predicting the therapeutic effect of $u_k$ on disease $v_m$ is essentially predicting the value of $\mathcal{A}_{km}$. $\mathcal{A} = \{\mathcal{A}^{train}, \mathcal{A}^{test} \}$, where $\mathcal{A}^{train}$ and $\mathcal{A}^{test}$ represent the training set and test set, respectively. The complete content of the proposed bidirectional behavioral sequence includes \textit{drug}, \textit{disease}, \textit{drug behavioral sequence}, \textit{disease behavioral sequence}, and \textit{label}. For the $\mathcal{A}_{km}$ prediction task, the drug and disease are denoted as $u_k$ and $v_m$, respectively. The drug behavior sequence is composed of diseases related to $u_k$ in the training set, excluding $v_m$. It can be defined as $\mathcal{R}^b_k = \{ v_j \mid 1 \leq j \leq |\mathcal{V}|, j \ne m, \mathcal{A}_{kj} \in \mathcal{A}^{train}\}$. The disease behavior sequence is composed of drugs related to $v_m$ in the training set, excluding $u_k$. It can be defined as $\mathcal{I}^b_m = \{ u_i \mid 1 \leq i \leq |\mathcal{U}|, i \ne k, \mathcal{A}_{im} \in \mathcal{A}^{train}\}$.  The label is $\mathcal{A}_{km}$. A complete behavior sequence is 
$(u_k, v_m, \mathcal{R}^b_k, \mathcal{I}^b_m, \mathcal{A}_{km}).$

\subsection{Constructing Latent Prototype Representations}\label{2.3}
In behavior sequence prediction tasks, the intrinsic attributes of users and items serve as vital characterizing information, enabling a more comprehensive understanding of entities. However, in drug repositioning tasks reliant on similarity metrics between entities, such internal attribute information remains underutilized. Consequently, we construct latent prototype representations for drugs and diseases using structural relationships derived from similarity data. These prototypes are structured into separate drug and disease prototype spaces to retain domain-specific semantics. As shown in Fig.~\ref{main}(a), initial representations from similarity matrices are refined via a Siamese-based network, and cosine similarity aligns the prototype spaces.

\subsubsection{Generating Initial Representations} 
The initial representation for each drug or disease is derived from its corresponding similarity matrix by isolating its full relational profile—formally equivalent to extracting either a row or column vector from the drug-drug ($\mathcal{S}^{\mathcal{U}}$) or disease-disease ($\mathcal{S}^{\mathcal{V}}$) similarity matrix, respectively. This vector extraction process preserves the complete similarity structure of each entity within its domain and provides geometrically interpretable input for subsequent prototype refinement. 

\subsubsection{Siamese Structures for Feature Learning} 
The initial similarity-based representations are processed through a Siamese network architecture~\cite{li2022survey} to refine their respective prototype vectors iteratively. The Siamese network comprises two parallel feature encoders with identical topological structures and shared weight parameters, ensuring parameter-space symmetry during joint optimization. This enhances the consistency of intra-domain feature transformation and reduces computational overhead compared to using separate encoders for feature extraction. 

\subsubsection{Prototype Alignment on Latent Space} 
After extracting prototype vectors, cosine similarity is used to align the entire prototype space, a metric invariant to vector magnitude that measures angular correspondence. The calculation method for the cosine similarity between prototypes is shown in Eq.~\eqref{eqcos}:
\begin{equation}\label{eqcos}
\mathrm{sim}(\mathbf{P}_{u_i}, \mathbf{P}_{u_j}) = \frac{\mathbf{P}_{u_i} \cdot \mathbf{P}_{u_j}}{\| \mathbf{P}_{u_i} \| \cdot \| \mathbf{P}_{u_j} \|}
\end{equation}
where $\mathbf{P}_{u_i}$ and $\mathbf{P}_{u_j}$ represent the output of the Siamase network for entities $u_i$ and $u_j$, respectively, and $\| . \|$ represents the $L^2$-norm. The angular relationships encoded by cosine similarity project the functional affinities between drugs and disease pairs onto a geometrically interpretable manifold in their respective high dimensional prototype spaces. By operating on direction rather than magnitude, this metric disentangles semantic relatedness from spurious scalar variations, effectively embedding drug and disease similarity as angular proximity in the latent space. To construct latent prototype spaces, the Siamese architectures are optimized via the contrastive loss defined in Eq.~\eqref{eqSim}. 
\begin{equation}\label{eqSim}
\begin{aligned}
\mathcal{L}_{\mathrm{sim}}^{\mathcal{U}}=\sum _{i=1}^{|\mathcal{U}|}\sum_{j=i+1}^{|\mathcal{U}|}\left( \mathcal{S}^{\mathcal{U}}_{ij}-\mathrm{sim}\left(f_\phi^{\mathcal{U}}(u_i), f_\phi^{\mathcal{U}}(u_j)\right)\right)^2 \\
\mathcal{L}_{\mathrm{sim}}^{\mathcal{V}}=\sum _{i=1}^{|\mathcal{V}|}\sum_{j=i+1}^{|\mathcal{V}|}\left( \mathcal{S}^\mathcal{V}_{ij}-\mathrm{sim}\left(f_\phi^{\mathcal{V}}(v_i), f_\phi^{\mathcal{V}}(v_j)\right)\right)^2
\end{aligned}
\end{equation}
where $f_\phi^{\mathcal{U}}$ and $f_\phi^{\mathcal{V}}$ represent the Siamese networks processing drug data and disease data, respectively. while $f_\phi^{\mathcal{U}}(u_i)$ and $f_\phi^{\mathcal{V}}(v_i)$ are equivalent to the extracted prototypes $\mathbf{P}_{u_i}$ and $\mathbf{P}_{v_i}$. This dual-stream refinement mechanism ensures that drug and disease prototypes capture domain-specific semantics with high fidelity while maintaining computational efficiency. The pre-trained drug and disease prototype encoders employing a Siamese architecture during this phase will be utilized in the following training stage to extract prototype representations for individual drugs and diseases. A comprehensive description of the methodological details governing this process is provided in Section~\ref{2.4}.

\subsection{Learning Bidirectional Behavioral Sequences}\label{2.4}
BiBLDR differs from conventional recommendation-based drug repositioning approaches by employing a Transformer architecture to learn bidirectional behavioral sequences. This design uses multi-head self-attention to model complex pharmacological interactions explicitly. At the same time, the Transformer's global context integration enables simultaneous capture of long-range intra-sequence relationships and inter-sequence dependencies. Besides, similarity information is further integrated at this stage. Fig.~\ref{main}(b) illustrates the main pipeline of the model. The methodological workflow can be systematically decomposed into four discrete computational phases: embedding projection, similarity information fusion, attention mechanism for sequence process, and drug-disease association prediction.

\subsubsection{Embedding Projection}
In addition to generating prototypes for drugs and diseases using similarity information in stage \Romann{1}, to further enrich the semantic information inherently embedded within drugs and diseases, through a structured embedding framework, each ID is mapped to a low-dimensional vector representation, termed an embedding, which is iteratively refined through optimization during model training. More precisely, two dedicated learnable matrices, $\mathcal{W}^\mathcal{U} \in \mathbb{R}^{|\mathcal{U}|\times d_w}$ and $\mathcal{W}^\mathcal{V} \in \mathbb{R}^{|\mathcal{V}|\times d_w}$ , are initialized to encode trainable embeddings for all $|\mathcal{U}|$ drugs and $|\mathcal{V}|$ diseases, respectively, where $d_w$ denotes the  dimensionality. These embeddings are incorporated into subsequent computational layers, enabling the model to capture relational properties through continual optimization, thereby distilling parametric relationships and latent interdependencies inherent in behavior sequence data.

\subsubsection{Similarity Information Fusion}
Although the entity prototypes derived from stage \Romann{1} training implicitly encapsulate similarity information, this remains insufficient. In this stage, we explicitly integrate the numerical similarity values between entities directly into the prototypes within the behavioral sequences. This integration amplifies the distinction in features among entities with varying degrees of similarity, rendering their differences more pronounced. For prediction task $\mathcal{A}_{km}$, the constructed bidirectional behavioral sequence consists of the drug-side behavior sequence $\mathcal{R}_k^b$ and the disease-side behavioral sequence $\mathcal{I}_m^b$. Initially, each drug and disease in $\mathcal{I}_m^b$ and $\mathcal{R}_k^b$, respectively, is embedded into the latent prototype space using the pre-trained prototype encoders from the Stage \Romann{1}, as shown in Eq.~\eqref{eqP}.
\begin{equation}\label{eqP}
\renewcommand{\arraystretch}{1.25}
\setlength{\arraycolsep}{1pt}
\begin{matrix}
 \mathcal{P}^\mathcal{U}_k  &  = &  \{ \mathbf{P}_{v_i} \mid \mathbf{P}_{v_i} = f^\mathcal{V}_\phi(v_i), v_i \in \mathcal{R}^b_k \} \\
\mathcal{P}^\mathcal{V}_m  &  = &  \{ \mathbf{P}_{u_i} \mid \mathbf{P}_{u_i} = f^\mathcal{U}_\phi(u_i), u_i \in \mathcal{I}^b_m \} 
\end{matrix}
\end{equation}
where $\mathbf{P}_{u_i}$ and $\mathbf{P}_{v_i}$ represent the extracted prototype from $u_i$ and $v_i$. To integrate entity similarity into prototypes, we convert numerical similarity scores between each entity in the behavioral sequence and the target entity into vectors matching the prototype’s dimensions. The padding method applied to these vectors is detailed in Eq.~\eqref{eqPAD}.
\begin{equation}\label{eqPAD}
\begin{matrix}
\mathbf{S}^\mathcal{U}_{ij} = \mathcal{S}^{\mathcal{U}}_{ij}\cdot \mathbf{1} \\
\mathbf{S}^\mathcal{V}_{ij} = \mathcal{S}^{\mathcal{V}}_{ij}\cdot \mathbf{1}
\end{matrix}
\end{equation}
where $\mathbf{S}^\mathcal{U}_{ij}$ represents the similarity vector generated between drug $u_i$ and $u_j$, $\mathbf{S}^\mathcal{V}_{ij}$ represents the similarity vector generated between disease $v_i$ and $v_j$, and $\mathbf{1}$ is a vector of ones with a dimension of $d_0$. To synthesize entity prototypes and their relational similarity context, the fusion layer combines the prototype with its corresponding similarity vector. This integration ensures that both intrinsic behavioral patterns and explicit similarity relationships are preserved in the feature representation. The merged feature vector is computed as shown in Eq.~\eqref{eqH}:
\begin{equation}\label{eqH}
\renewcommand{\arraystretch}{1.25}
\setlength{\arraycolsep}{1pt}
\begin{matrix}
\mathcal{H}^{\mathcal{U}}_{km}  & = & \{ \mathbf{H}^\mathcal{U}_{mi} \mid \mathbf{H}^{\mathcal{U}}_{mi} & = & f_\varphi ^\mathcal{U}
 (\mathbf{P}_{v_i}\oplus\mathbf{S}^{\mathcal{V}}_{mi}), \mathbf{P}_{v_i} \in \mathcal{P}^\mathcal{U}_k\}\\
\mathcal{H}^{\mathcal{V}}_{km}  & = & \{ \mathbf{H}^\mathcal{V}_{ki} \mid  \mathbf{H}^{\mathcal{V}}_{ki} & = & f_\varphi ^\mathcal{V}
  (\mathbf{P}_{u_i}\oplus\mathbf{S}^{\mathcal{U}}_{ki}), \mathbf{P}_{u_i} \in \mathcal{P}^\mathcal{V}_m\}
\end{matrix}
\end{equation}
where $\oplus$ represents the feature concatenation operation, $f_\varphi^\mathcal{U}$ and $f_\varphi ^\mathcal{V}$ are the fusion layers on drug-side and disease-side, respectively. In our model, the fusion layer comprises a feedforward neural network. $\mathbf{H}^{\mathcal{U}}_{ki}$ and $\mathbf{H}^{\mathcal{V}}_{mj}$ are the fused features, with a dimensional size of $d_0$. 

\subsubsection{Attention Mechanism for Sequence Process} 
\begin{figure}[htbp]
    \centering
    \includegraphics[width=0.45\linewidth]{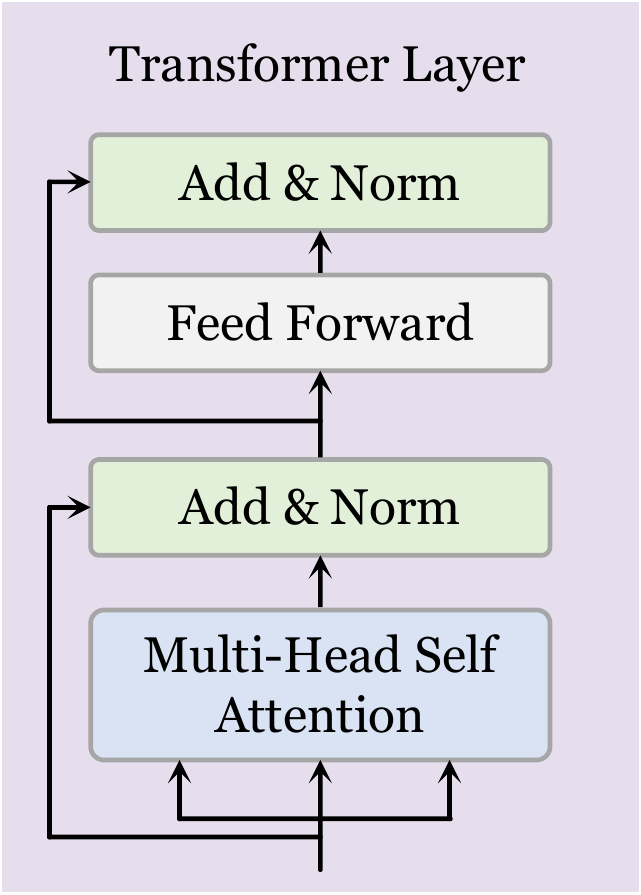}
    \caption{The structure of the Transformer layer includes the multi-head self-attention mechanism, normalization layer, feedforward neural network, and residual connections.}
    \label{TF}
\end{figure}

Following the fusion of similarity information into the prototypes of entities within bidirectional behavioral sequences, the entity embeddings from $\mathcal{W^U}$ and $\mathcal{W^V}$ are concatenated with similarity-fused feature vectors to enhance representational capacity, while rating in behavior sequence is also incorporated. In contrast to conventional approaches that directly multiply feature vectors and ratings, our approach introduces a logarithmic transformation to the binarized rating values prior to their interaction with the feature vectors. This operational approach amplifies the influence of positive samples within behavioral sequences on the drug repositioning process while simultaneously retaining the semantic integrity of negative samples to facilitate contrastive learning in the model. Eq.~\eqref{eqI} formalizes this operation.
\begin{equation}\label{eqI}
\renewcommand{\arraystretch}{1.25}
\setlength{\arraycolsep}{1pt}
\begin{matrix}
\mathcal{N}^{\mathcal{U}}_{km} &=& \{ \mathbf{X}^\mathcal{U}_{mi} \mid \mathbf{X}^\mathcal{U}_{mi} = \left(\mathbf{H}^\mathcal{U}_{mi} \oplus \mathcal{W}^\mathcal{V}_i\right) e^{\mathcal{T}\cdot\mathcal{A}_{ki}}, \mathbf{H}^\mathcal{U}_{mi} \in \mathcal{H}^{\mathcal{U}}_{km}  \} \\
\mathcal{N}^{\mathcal{V}}_{km} &=& \{ \mathbf{X}^\mathcal{V}_{ki} \mid \mathbf{X}^\mathcal{V}_{ki} = \left(\mathbf{H}^\mathcal{V}_{ki} \oplus \mathcal{W}^\mathcal{U}_i\right)e^{\mathcal{T}\cdot\mathcal{A}_{im}}, \mathbf{H}^\mathcal{V}_{ki} \in \mathcal{H}^{\mathcal{V}}_{km}  \}
\end{matrix}
\end{equation}
where $\mathcal{T}$ is a tunable hyperparameter, and $\mathcal{W}^\mathcal{U}_i$ and $\mathcal{W}^\mathcal{V}_i$ represent the $i$-th row of $\mathcal{W}^\mathcal{U}$ and the $i$-th row of $\mathcal{W}^\mathcal{U}$, which correspond to the embedding of the $i$-th disease and the embedding of the $i$-th drug, respectively. $\mathcal{N}^{\mathcal{U}}_{km}$ and $\mathcal{N}^{\mathcal{V}}_{km}$ represent the sequence sets on the drug-side and disease-side, respectively, which are the inputs to the Transformer layer. Following~\cite{chen2019behavior}, we use a Transformer layer based on a multi-head self-attention mechanism to process the bidirectional behavioral sequence features $\mathcal{N}^{\mathcal{U}}_{km}$ and $\mathcal{N}^{\mathcal{V}}_{km}$. For ease of understanding, $\mathcal{N}_{km} = \{ \mathcal{N}_{km}^\mathcal{U}, \mathcal{N}_{km}^\mathcal{V} \}$ is simplified as $\mathbf{X} \in \mathbb{R}^{L_{\mathrm{seq}}\times d_1}$ when describing the forward propagation process of the Transformer layer. $d_1 = d_w + d_0$, $L_{\mathrm{seq}}$ represents the aggregate count of all sequences contained within $\mathcal{N}^{\mathcal{U}}_{km}$ and $\mathcal{N}^{\mathcal{V}}_{km}$. The structural diagram of the Transformer layer is shown in Fig.~\ref{TF}.

A Transformer layer has a total of $n$ attention heads, the input matrix $\mathbf{X}$ will be projected into $n$ different low-dimensional subspaces by multiplying learnable weight matrices and then scaled dot-product attention will be computed, as shown in Eq.~\eqref{eqHead}.
\begin{equation}\label{eqHead}
\begin{aligned}
\mathrm{head}_i = \operatorname{Attention}(\mathbf{X}\mathbf{W}_i^Q, \mathbf{X}\mathbf{W}_i^K, \mathbf{X}\mathbf{W}_i^V) \\
\mathrm{Attention}\left(\mathbf{Q}_{i}, \mathbf{K}_{i}, \mathbf{V}_{i}\right)=\operatorname{softmax}\left(\frac{\mathbf{Q}_{i} \mathbf{K}_{i}^{T}}{\sqrt{d_{k}}}\right) \mathbf{V}_{i}
\end{aligned}
\end{equation}
where $\mathbf{Q}_i$ represents the queries of the $i$-head, $\mathbf{K}_i$ represents the keys, and $\mathbf{V}_i$ represents the values, these values are derived by computing the product of $\mathbf{X}$ and each respective learned weight matrix ($\mathbf{W}_i^Q, \mathbf{W}_i^K, \mathbf{W}_i^V \in \mathbb{R}^{d_1 \times d_k}$). $d_k = d_1 / n$, $\sqrt{d_k}$ is used to scale the scores to mitigate gradient vanishing. The outputs from all attention heads are concatenated and subsequently multiplied by a trainable weight matrix, as shown in Eq.~\eqref{eqMH}.
\begin{equation}\label{eqMH}
\mathrm{MH}(\mathbf{X}) = \mathrm{Concat}(\mathrm{head}_1, \cdots, \mathrm{head}_n)\cdot\mathrm{W}^O
\end{equation}
where $\mathbf{W}^O$ is a trainable weight matrix, $\mathrm{Concat}$ denotes the feature concation operation, and $\mathrm{MH}(\mathbf{X})$ represents the output of the multi-head self-attention mechanism. The output of subsequent layers are shown in Eq~\eqref{eqO}.
\begin{equation}\label{eqO}
\begin{aligned}
\mathbf{O} = \mathrm{BN}(\mathbf{X}+\mathrm{MH}(\mathbf{X})) \\
\mathbf{O'} = \mathrm{BN}(\mathbf{O} + \mathrm{FFN}(\mathbf{O}))
\end{aligned}
\end{equation}
where $\mathrm{BN}$ denotes the batch normalization layer, $\mathrm{FFN}$ represents a feed-forward neural network responsible for endowing the Transformer layer with nonlinear transformation capability. $\mathbf{O'}$ is the final output of the Transformer layer, which can be expressed as $\mathcal{O}_{km}^{\mathrm{TL}} \in \mathbb{R}^{L_{\mathrm{seq}} \times d_1 }$

\subsubsection{Drug-disease Association Prediction}
In addition to the bilateral behavioral sequence features, we integrate the intrinsic attributes of the drug $u_k$ and the disease $v_m$ into the final prediction of the association $\mathrm{A}_{km}$. Specifically, we extract the prototypes and embeddings of $u_k$ and $v_m$, which are then concatenated with the bilateral behavioral sequence features $\mathcal{O}_{km}^{\mathrm{TL}}$ to form the input for the association prediction model. To ensure semantic consistency, the prototypes and embeddings are first processed through an alignment layer, which harmonizes their semantic information with the output of the Transformer layer before concatenation with $\mathcal{O}_{km}^{\mathrm{TL}}$. This progress is shown in Eq.~\eqref{eqM}.
\begin{equation}\label{eqM}
\mathbf{M}_{km} = f_\alpha^\mathcal{U}(\mathbf{P}_{u_k}\oplus\mathcal{W}^{\mathcal{U}}_k)\oplus\mathcal{O}_{km}^{\mathrm{TL}} \oplus f^\mathcal{V}_{\alpha}(\mathbf{P}_{v_m} \oplus \mathcal{W}^{\mathcal{V}}_m)
\end{equation}
where $f_\alpha^\mathcal{U}$ and $f_\alpha^\mathcal{V}$ represent the alignment layers for the drug and disease sides, respectively, each constructed using feedforward neural networks. Prior to concatenation, $\mathcal{O}_{km}^{\mathrm{TL}}$ is also flattened into a one-dimensional vector. The association prediction model processes a condensed one-dimensional feature vector, representing comprehensively encoded drug-disease interaction patterns, to generate a scalar output quantifying the predicted therapeutic relationship. This component is architecturally implemented as a multilayer feedforward neural network characterized by successive non-linear transformations. During the optimization phase of Stage II, a task-specific binary cross-entropy loss function is implemented to facilitate the joint optimization of all model components. This loss function is specifically tailored for binary classification tasks and leverages backpropagation to iteratively refine the network's trainable parameters, thereby ensuring comprehensive convergence of the integrated architecture. The loss function is shown in Eq.~\eqref{eqLBCEWL}.
\begin{equation}\label{eqLBCEWL}
\begin{aligned}
\mathcal{L}_{BCEWL}=& \frac{1}{N} \sum_{i=1}^{N}(y_{i} \cdot \log (\sigma(p_{i}))\\
&+(1-y_{i}) \cdot \log (1-\sigma(p_{i}))) \\
\end{aligned}
\end{equation}
where $N$ represents the total number of training samples, $\sigma$ denotes the Sigmoid function, each sample $x_i$ has a binary label $y_i \in \{ 0, 1 \}$, and $p_i$ represents the predicted probability of sample $x_i$ being classified as positive. In summary, the comprehensive workflow of our proposed bidirectional behavioral sequence learning strategy is formally outlined in Algorithm~\ref{A1}.

\begin{algorithm}[h]\label{A1}
\caption{Bidirectional behavior learning on BiBLDR}
\label{a}
\KwIn{test dataset $\mathcal{A}^{test}$, promote matrix $\mathcal{W}^\mathcal{U}$ and $\mathcal{W}^\mathcal{V}$, prototype encoder $f_\phi^\mathcal{U}$ and $f_\phi^\mathcal{V}$, trained model (fusion layer $f_\varphi^\mathcal{U}$ and $f_\varphi^\mathcal{V}$, Transformer layer $\tau$, align layer $f_\alpha^\mathcal{U}$ and $f_\alpha^\mathcal{V}$, and association prediction layer $f_\delta$).}
\KwOut{Predicted $\mathcal{A}^{test}$}
\For{$\mathcal{A}_{km} \in \mathcal{A}^{test}$}{
    \codezs{Drug-side sequence process}
     $\mathcal{R}_k^b \leftarrow$ Get drug-side behavior sequence of $u_k$\;
     \For{$v_i \in \mathcal{R}_k^b$}{
        $\mathbf{P}_{v_i} \leftarrow f_\phi^\mathcal{V}(v_i)$\;
        $\mathbf{S}_{mi}^\mathcal{V} \leftarrow \mathcal{S}_{mi}^{\mathcal{V} }\cdot \mathbf{1} $\;  
        $\mathbf{H}_{mi}^\mathcal{U} \leftarrow f_\varphi ^\mathcal{U}
 (\mathbf{P}_{v_i}\oplus\mathbf{S}^{\mathcal{V}}_{mi}) $\;
        $\mathbf{X}_{mi}^{\mathcal{U}} \leftarrow \left(\mathbf{H}^\mathcal{U}_{mi} \oplus \mathcal{W}^\mathcal{V}_i\right) e^{\mathcal{T}\cdot\mathcal{A}_{ki}} $\;
        $\mathcal{N}_{km}^{\mathcal{U}}$ append $\mathbf{X}_{mi}^{\mathcal{U}}$\;
     }
     \codezs{Disease-side sequence process}
    $\mathcal{I}_k^b \leftarrow$ Get disease-side behavior sequence of $v_m$\;
     \For{$u_i \in \mathcal{I}_k^b$}{
        $\mathbf{P}_{u_i} \leftarrow f_\phi^{\mathcal{U}}(u_i)$\;
        $\mathbf{S}_{ki}^{\mathcal{U}} \leftarrow \mathcal{S}_{ki}^\mathcal{U}\cdot\mathbf{1}$\;
        $\mathbf{H}_{ki}^{\mathcal{V}} \leftarrow f_\varphi ^\mathcal{V}
  (\mathbf{P}_{u_i}\oplus\mathbf{S}^{\mathcal{U}}_{ki})$\;
        $\mathbf{X}_{ki}^{\mathcal{U}} \leftarrow \left(\mathbf{H}^\mathcal{V}_{ki} \oplus \mathcal{W}^\mathcal{U}_i\right)e^{\mathcal{T}\cdot\mathcal{A}_{im}}$\;
        $\mathcal{N}_{km}^{\mathcal{V}}$ append $\mathbf{X}_{ki}^{\mathcal{V}}$\;
     }
     $\mathcal{N}_{km} \leftarrow \{ \mathcal{N}_{km}^{\mathcal{U}}, \mathcal{N}_{km}^{\mathcal{V}}\}$\;
     $\mathcal{O}_{km}^{\mathrm{TL}}\leftarrow \tau(\mathcal{N}_{km})$\;
     $\mathbf{M}_{km} \leftarrow f_\alpha^\mathcal{U}(\mathbf{P}_{u_k}\oplus\mathcal{W}^{\mathcal{U}}_k)\oplus\mathcal{O}_{km}^{\mathrm{TL}} \oplus f^\mathcal{V}_{\alpha}(\mathbf{P}_{v_m} \oplus \mathcal{W}^{\mathcal{V}}_m)$\;
     $\mathcal{A}_{km} \leftarrow f_\delta(\mathbf{M}_{km})$
}
\end{algorithm}
 
\section{Experiments And Discussion}\label{Section3}
In this section, We first describe the dataset used, the evaluation metrics, and the experimental implementation details. Next, We perform 10-fold cross-validation experiments using BiBLDR with other baseline models on the benchmark datasets to validate the performance achieved by BiBLDR. Ablation experiments are also configured to verify the contribution of individual components in BiBLDR. Additionally, we conduct cold-start experiments, sparse environment testing, and parameter analysis. These experiments validate BiBLDR's capability in identifying new drug indications, adaptability to sparse data, and the impact of hyperparameter settings on performance. Finally, we conduct case studies to validate the practical applicability of BiBLDR.

\subsection{Datasets and Evaluation Metrics}
To comprehensively evaluate BiBLDR's performance, we tested it on three benchmark datasets: Gdataset~\cite{gottlieb2011predict}, Cdataset~\cite{luo2016drug}, and LRSSL~\cite{liang2017lrssl}. The Gdataset comprises 593 drugs sourced from DrugBank~\cite{wishart2006drugbank} and 313 diseases from \emph{Online Mendelian Inheritance in Man} (OMIM)~\cite{hamosh2005online}, with 1,933 experimentally validated drug-disease associations. The Cdataset includes 633 drugs and 409 diseases, featuring 2,352 validated drug-disease associations curated from the Comparative Toxicogenomics Database~\cite{davis2017comparative}. Lastly, the LRSSL dataset contains 763 drugs and 681 diseases, encompassing 3,051 known drug-disease interactions. Following~\cite{sun2022partner, sun2024drug}, we derive drug similarity matrices using 2D chemical fingerprints and disease similarity matrices using phenotypic features. These matrices capture structural and functional relationships between drugs and diseases, respectively, and are integrated as auxiliary data during training. Tab.~\ref{dataset} provides an overview of the key characteristics and statistical details for these datasets. We select \emph{Area Under the Precision-Recall Curve} (AUPRC)  and  \emph{Area Under the Receiver Operating Characteristic Curve} (AUROC) as evaluation metrics to assess model performance. These metrics are widely adopted in drug repositioning tasks due to their robustness and relevance.
\begin{table}[htbp]
    \centering
    \renewcommand{\arraystretch}{1.25}
    \caption{Statistic information about benchmark datasets. \textbf{No. of drugs}, \textbf{No. of diseases}, and \textbf{No. of associations} represent the total number of drugs, diseases, and known drug-disease associations, respectively. \textbf{Sparsity} represents the degree of sparsity in the drug-disease association matrix.}
    \resizebox{0.49\textwidth}{!}{
    \begin{tabular}{lcccc}
    \toprule[1.5pt]
        \textbf{Dataset} & \textbf{No. of drugs}& \textbf{No. of diseases} & \textbf{No. of associations} & \textbf{Sparsity} \\
    \hline
     Gdataset  & 593 & 313 & 1933 & 0.0104\\
     Cdataset & 663 & 409 & 2532 & 0.0093\\
     LRRSL & 269 & 598 & 18416 & 0.1145\\
     \bottomrule[1.5pt]
    \end{tabular}}
    \label{dataset}
\end{table}

\subsection{Implementation Details}
To ensure the rigor of performance evaluation, we report model performance using 10-fold cross-validation. We first divide all positive samples in the drug-disease association matrix $\mathcal{A}$ into 10 equal folds $\{ \mathcal{A}^{\mathrm{pos}}_1, \mathcal{A}^{\mathrm{pos}}_2, \cdots, \mathcal{A}^{\mathrm{pos}}_{10} \}$. An equal number of negative samples are randomly selected from $\mathcal{A}$ and split into 10 corresponding folds $\{ \mathcal{A}^{\mathrm{neg}}_1, \mathcal{A}^{\mathrm{neg}}_2, \cdots, \mathcal{A}^{\mathrm{neg}}_{10} \}$. These folds are paired to create 10 cross-validation subsets $\mathcal{T}_{\mathrm{cv}} = \{ (\mathcal{A}^{\mathrm{pos}}_i, \mathcal{A}^{\mathrm{neg}}_i) \mid 1 \leq i \leq 10 \}$. During evaluation, we rotate one fold as the test set and use the remaining nine for training, repeating this until all folds have served as the test set. To ensure statistical reliability, the experiment is repeated 10 times independently, with final results averaged across all runs. We employ AdamW as the optimizer for the neural network and utilize a cosine annealing scheduler for learning rate adjustment. During stage \Romann{1}, the learning rate is set to 0.01, while in stage \Romann{2}, it is reduced to 0.0001. The implementation is developed using Python 3.8.19 and PyTorch 2.4.0, and all experiments are conducted on an NVIDIA GeForce RTX 4070 Laptop GPU.

\subsection{Comparative Analysis with Baselines}
To validate BiBLDR’s state-of-the-art performance, we benchmark it against diverse baseline models on benchmark datasets. The baselines encompass non-deep learning methods: SCMFDD~\cite{zhang2018predicting}, iDrug~\cite{chen2020idrug}, BNNR~\cite{yang2019drug}, and NRLMF~\cite{liu2016neighborhood} and deep learning approaches: NIMCGAN~\cite{li2020neural}, DRWBNCF~\cite{meng2022weighted}, and PSGCN~\cite{sun2022partner}. This selection ensures robust validation across traditional and modern drug repositioning methodologies. Detailed performance metrics for all models are provided in Tab.~\ref{Tab10}. The experimental results demonstrate that our proposed BiBLDR performs optimally in AUROC and AUPRC metrics. BiBLDR achieves remarkable AUROC and AUPRC of 0.9978 and 0.9982 on the Cdataset, outperforming the second-ranked by margins of 0.0412 and 0.0328. On the additional benchmark datasets, Gdataset and LRSSL, BiBLDR maintains its dominant performance, achieving AUROC of 0.9941 and 0.9950 and AUPRC of 0.9950 and 0.9964, respectively. BiBLDR demonstrates significant superiority over traditional non-deep-learning-based drug repositioning methods across all three benchmark datasets. Moreover, it also exhibits a clear advantage compared to current state-of-the-art methods based on deep learning and GCNs.

\begin{table*}[htbp]
    \caption{A comparative experiment involving 10 times 10-fold cross-validation with other models on Gdataset, Cdataset, and LRSSL. The highest-performing metric is denoted in bold.}
    \label{Tab10}
    \centering
    \renewcommand{\arraystretch}{1.25}
    \begin{tabular}{cc|cccc|ccccc}
    \toprule[1.5pt]
        \multirow{2}{*}{\textbf{Metric}} & \multirow{2}{*}{\textbf{Dataset}} & \multicolumn{4}{c|}{\textbf{Non-Deep Learning-Based Methods}} & \multicolumn{4}{c}{\textbf{Deep Learning-Based Methods}}   \\
        \cline{3-10}
        ~ & ~ & \textbf{SCMFDD} & \textbf{iDrug} & \textbf{BNNR} & \textbf{NRLMF} & \textbf{NIMCGCN} & \textbf{DRWBNCF} & \textbf{PSGCN} & \textbf{BiBLDR (Ours)} \\
        \hline
        \multirow{3}{*}{AUROC} & Gdataset & 0.7731 & 0.9078 & 0.9412 & 0.9097 & 0.8234 & 0.9061 & 0.9485 & \textbf{0.9941} \\
        ~ & Cdataset & 0.7896 & 0.9294 & 0.9522 & 0.9257 & 0.8393 & 0.9277 & 0.9566 & \textbf{0.9978} \\
        ~ & LRSSL & 0.7698 & 0.8993 & 0.9201 & 0.8854 & 0.7581 & 0.9232 & 0.9395 & \textbf{0.9950} \\
        \hline
        \multirow{3}{*}{AUPRC} & Gdataset & 0.7749 & 0.9265 & 0.9575 & 0.9302 & 0.8590 & 0.9307 & 0.9558 & \textbf{0.9950} \\
        ~ & Cdataset & 0.7878 & 0.9454 & 0.9654 & 0.9441 & 0.8728 & 0.9476 & 0.9627 & \textbf{0.9982} \\
        ~ & LRSSL & 0.7860 & 0.9212 & 0.9434 & 0.9102 & 0.7962 & 0.9339 & 0.9462 & \textbf{0.9964} \\
        \bottomrule[1.5pt]
    \end{tabular}
\end{table*}

\subsection{Ablation Studies}
To assess the contribution of individual modules in BiBLDR, we perform ablation studies on its components. The detailed results are summarized in Tab.~\ref{Ablation}. [A] signifies the exclusion of the entity's prototype, relying solely on the embedding vector to represent the entity's attributes. The results show that the prototype space for drugs and diseases enhances the intrinsic attributes of entities. [B] and [C] represent the use of only disease-side and drug-side behavioral sequence data, respectively, for model training. The results show that the loss of either side in the bidirectional behavioral sequences weakens the representational capacity of the behavioral sequences, leading to a decline in model performance. Particularly, when disease-side data is missing, AUROC and AUPRC experience a sharp decline. [D] demonstrates that the similarity information fusion strategy in stage \Romann{2} effectively enhances the representational capability of prototypes. [E] illustrates that applying the multi-head self-attention mechanism enables BiBLDL to capture the interactive relationships among multiple behavioral sequence features more accurately.

\begin{table*}[htbp]
    \caption{Ablation experiments of BiBLDR on Gdataset. \textbf{PSC} represents prototype space construction. \textbf{R-S} and \textbf{D-S} represents drug and disease behavior sequence data, respectively. \textbf{SF} represents similarity fusion in stage \Romann{2}. \textbf{MF} represents the application of multi-head self-attention.}
    \label{Ablation}
    \renewcommand{\arraystretch}{1.25}
    \centering
    \begin{tabular}{c|ccccc|cccccc}
    \toprule[1.5pt]
        ~ & \multicolumn{5}{c|}{\textbf{Ablation Components}} & \multicolumn{2}{c}{\textbf{Gdataset}} & \multicolumn{2}{c}{\textbf{Cdataset}} & \multicolumn{2}{c}{\textbf{LRSSL}}\\
        \cline{2-12}
        ~ &\textbf{PSC} & \textbf{R-S} & \textbf{D-S} & \textbf{SF} & \textbf{MA} & \textbf{AUROC} & \textbf{AUPRC} & \textbf{AUROC} & \textbf{AUPRC} & \textbf{AUROC} & \textbf{AUPRC} \\
        \hline
        [A] & \ding{55} & \ding{51} & \ding{51} & \ding{51} & \ding{51} & 0.9597 & 0.9678 & 0.9766 & 0.9815 & 0.9031 & 0.9269  \\
        
        [B] &\ding{51} & \ding{55} & \ding{51} & \ding{51} & \ding{51} &0.9674 & 0.9724 & 0.9798 & 0.9849 & 0.9673 & 0.9774  \\
        
        [C] & \ding{51} & \ding{51} & \ding{55} & \ding{51} & \ding{51} & 0.9131 & 0.9289 & 0.9537 & 0.9606 & 0.8011 & 0.8328 \\
        
        [D] & \ding{51} & \ding{51} & \ding{51} & \ding{55} & \ding{51} & 0.9645 & 0.9616  & 0.9769 & 0.9781 & 0.9533 & 0.9605 \\
        
        [E] & \ding{51} & \ding{51} & \ding{51} & \ding{51} & \ding{55} & 0.9889 & 0.9925 & 0.9906 & 0.9940 & 0.9903 & 0.9932 \\
        
        [F] & \ding{51} & \ding{51} & \ding{51} & \ding{51} & \ding{51} & 0.9941 & 0.9950 & 0.9978 & 0.9982 & 0.9950 & 0.9964\\
         \bottomrule[1.5pt]
    \end{tabular}
\end{table*}

\subsection{Discovering Candidates for New Drugs}
\begin{figure}[htbp]
    \centering
    \includegraphics[width=\linewidth]{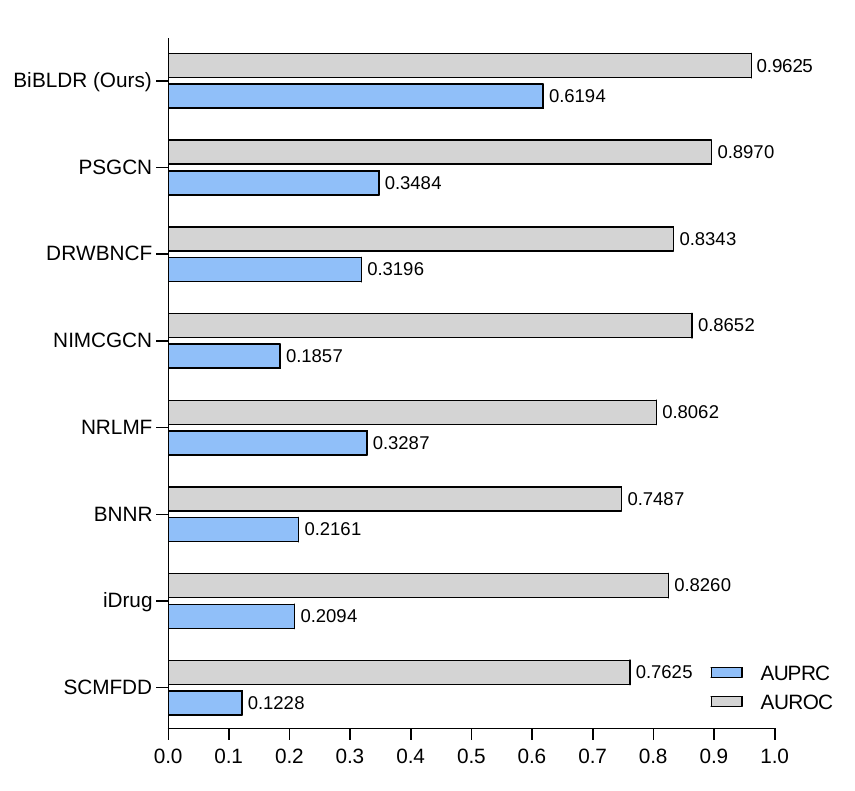}
    \caption{Comparison of cold-start experimental performance of different methods on the Gdataset.}
    \label{denovo}
\end{figure}

To assess BiBLDR's capability to discover therapeutic indications for novel drugs, we perform a cold-start experiment using training data. Here, new drugs are defined as those lacking known disease associations. For the $i-$th drug $u_i$, we mask all its disease associations in $\mathcal{A}$, reserving these as the test set. The remaining associations in $\mathcal{A}$ are the training set. This setup evaluates BiBLDR's ability to predict therapeutic effects of $u_i$ across diseases when no prior associations exist, forcing the model to rely exclusively on disease-side behavioral sequences. We test BiBLDR on Gdataset for indication discovery. As shown in Fig.~\ref{denovo}, BiBLDR performs best on both metrics. Notably, it demonstrates a significant and overwhelming improvement in AUPRC, reaching 0.6194, while the second-best result is only 0.3484. The AUROC also reaches an impressive 0.9625. These cold-start experiments fully demonstrate the remarkable application value of BiBLDR in novel drug discovery.

\subsection{Sparse Environment Testing}
\begin{figure}[htbp]
    \centering
    \includegraphics[width=1\linewidth]{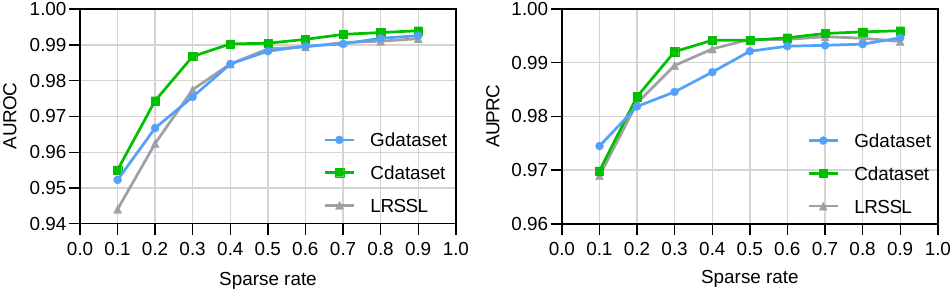}
    \caption{Sparse environment testing result on Gdataset, Cdataset, and LRSSL.}
    \label{SET}
\end{figure}

Drug repositioning tasks frequently involve managing highly sparse datasets in practical settings. To assess the robustness of BiBLDR under sparse data conditions, we implement a sparse parameter $\lambda$, which denotes the fraction of known associations sampled from the original dataset for model training. This configuration establishes a sparse, class-imbalanced data environment. Experiments are conducted by varying lambda across the range of $\{10\%, 20\%, 30\%, 40\%, 50\%, 60\%, 70\%, 80\%, 90\% \}$, and the corresponding results are depicted in the accompanying Fig.~\ref{SET}.The results reveal that even under the extreme sparsity condition of $\lambda = 10\%$, the AUROC values for Gdataset and Cdataset consistently exceed 0.95, while LRSSL achieves a value above 0.94. Furthermore, Gdataset's AUPRC surpasses 0.97, with the other two datasets maintaining values close to 0.97. Notably, when the known drug-disease associations available for training are extremely sparse, the effective information carried by the behavioral sequences constructed from the association matrix also diminishes accordingly. The reason BiBLDR can still achieve robust performance when the bahevioral sequence representation are insufficient lies in the fact that drug and disease prototypes are also incorporated as integral components of the association prediction. The prototype encoders are thoroughly pre-trained in the first stage, independent of the sparsity of known associations. This experiment demonstrates that using cosine similarity in the feature space to model the similarity between entities is reliable. Moreover, having semantically rich representations provides a fundamental performance guarantee for drug–disease association prediction when behavioral sequences are sparse. These findings highlight BiBLDR's exceptional ability to handle sparsity, demonstrating robust performance even in scenarios with limited sample availability.

\subsection{Parameter Analysis}
\begin{figure}[htbp]
    \centering
    \includegraphics[width=1.0\linewidth]{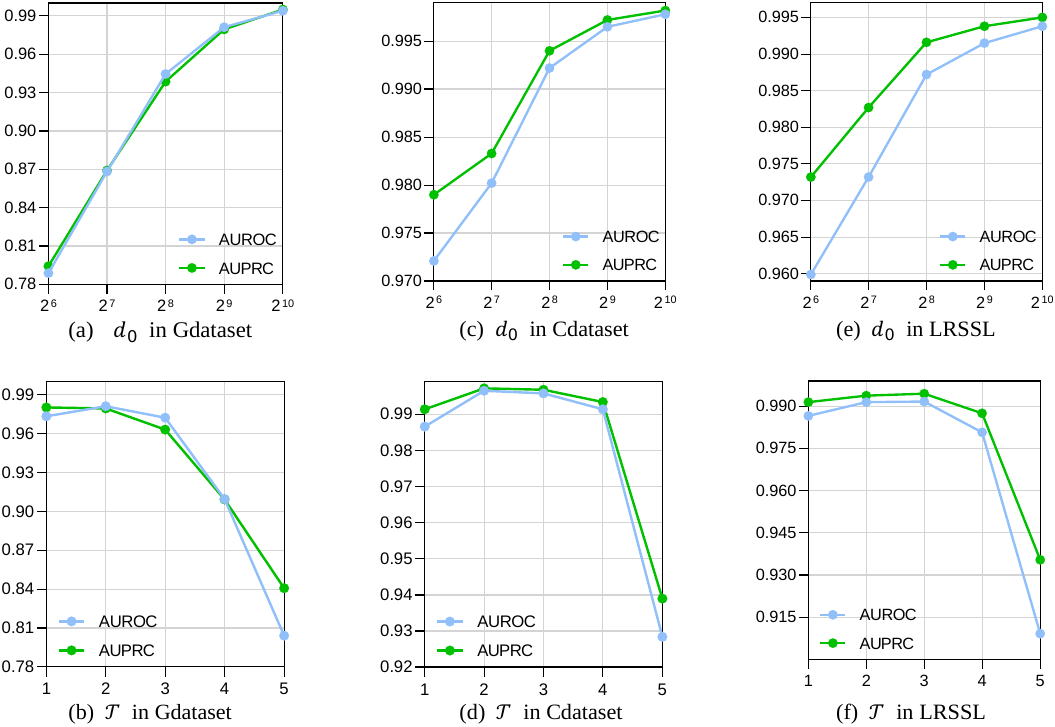}
    \caption{Parameter anlysis experiments on Gdataset, Cdataset, and LRSSL.}
    \label{PA}
\end{figure}

We further investigate the impact of parameters $d_0$ and $\mathcal{T}$ on the experimental performance. $d_0$ represents the vector length in the prototype space, which is utilized to shape the prototype vectors. $\mathcal{T}$ serves as the exponential temperature coefficient when integrating the scoring sequence, enabling the differentiation between positive and negative samples within the behavioral sequence. We sample $d_0$ in the range of $\{2^6, 2^7, 2^8, 2^9, 2^{10}\}$, $\mathcal{T}$ in the range of $\{1, 2, 3, 4, 5\}$. We conduct experiments using these parameters on the benchmark datasets, and the results are shown in Fig.~\ref{PA}. For $d_0$, as the values increase, performance shows an upward trend on all datasets, with $2^{10}$ being the optimal choice. This indicates that higher-dimensional prototypes can represent richer information. When $\mathcal{T} > 3$, as the values increase, model performance shows a significant downward trend. This indicates that when the numerical gap between positive and negative samples becomes too large, the model struggles to balance the representation learning of both positive and negative samples. For Gdataset, Cdataset, and LRSSL, the optimal $\mathcal{T}$ parameters are $\{2, 2, 3\}$.

\subsection{Case Studies}
To evaluate the real-world applicability of BiBLDR in drug repositioning, we perform case studies focusing on two diseases: lung cancer and hypertension. Specifically, we utilize all known drug-disease associations in Gdataset as the training set to train the model. Subsequently, all rest associations with lung cancer and hypertension are treated as candidate pairs, and the trained model is used to predict scores for each candidate pair. For both lung cancer and hypertension, the top-10 candidate drugs are ranked according to their predicted scores, and each was individually validated to confirm the existence of a therapeutic relationship. 

\subsubsection{Lung Cancer (211980 in OMIM)} Lung cancer is one of the cancers with the highest incidence and mortality rates worldwide, feared for its high fatality rate and lack of noticeable early symptoms. The left of Tab.~\ref{CS} provides the top-10 predicted candidate drugs for lung cancer, as determined by BiBLDR. Eight of the 10 candidate lung cancer drugs are validated as accurate predictions. For instance, combination chemotherapy regimens involving cisplatin or carboplatin with etoposide have demonstrated effectiveness in treating both non-small and small cell lung cancer~\cite{jang2025etoposide}. Etoposide is the highest-ranked candidate drug in Tab.~\ref{CS}.

\subsubsection{Hypertension (145500 in OMIM)} Hypertension is a prevalent and highly detrimental chronic condition, known for its subtle onset and multifaceted causes. It is especially common among middle-aged and elderly individuals, though it can also occur in younger populations. The right of Tab.~\ref{CS} provides the top-10 predicted candidate drugs for hypertension. nine of the top-ranked candidate drugs predicted by BiBLDR have been supported by valid evidence, achieving a 90\% hit rate. For instance, methyclothiazide has been proven effective in treating hypertension caused by heart failure, kidney failure, or estrogen and steroid therapy. Methyclothiazide is the tenth-ranked candidate drug in Tab.~\ref{CS}.

\begin{table*}[htbp]
    \caption{Top-10 candidate drug of lung cancer and hypertension. Left is lung cancer and right is hypertension.}
    \label{CS}
    \centering
    \renewcommand{\arraystretch}{1.25}
    \begin{tabular}{c|ccc|ccc}
    \toprule[1.5pt]
    \multirow{2}{*}{\textbf{Rank}} & \multicolumn{3}{c|}{\textbf{Lung Cancer}} & \multicolumn{3}{c}{\textbf{Hypertension}} \\
    \cline{2-7}
    ~ &\textbf{ DrugBank ID} & \textbf{Candidate Drug} & \textbf{Evident} & \textbf{DrugBank ID} & \textbf{Candidate Drug} & \textbf{Evident} \\
    \hline
      1   & DB00773 & Etoposide & \cite{jang2025etoposide} &
      DB01023 & Felodipine & \cite{mace1985felodipine}  \\
      2 & DB01030 & Topotecan & \cite{horita2015topotecan} & 
      DB01244 & Bepridil & \cite{aoki1986hypotensive} \\
      3 & DB00563 & Methotrexate & \cite{smyth1981methotrexate} &
      DB00310 & Chlorthalidone & \cite{toubes1971hypotensive} \\
      4 & DB00232 & Methyclothiazide & \textbf{-} & 
      DB01611 & Hydroxychloroquine & \cite{choi2022hydroxychloroquine} \\
      5 & DB00762 & Irinotecan & \cite{sevinc2011irinotecan} & 
      DB00808 & Indapamide & \cite{kaplan2015indapamide} \\
      6 & DB00853 & Temozolomide & \cite{dziadziuszko2003temozolomide} &
      DB00800 & Fenoldopam & \cite{murphy2001fenoldopam} \\
      7 & DB01268 & Sunitinib & \cite{tanday2015sunitinib} & 
      DB00091 & Cyclosporine & \textbf{-} \\
      8 & DB00694 & Daunorubicin & \cite{wang2019inhibition} & 
      DB00178 & Ramipril & \cite{frampton1995ramipril} \\
      9 & DB00859 & Penicillamine & \cite{SCIEGIENKA2017354} & 
      DB00880 & Chlorothiazide & \cite{freis1958chlorothiazide}\\
      10 & DB00635 & Prednisone & \textbf{-} & 
      DB00232 & Methyclothiazide & \cite{black1992antihypertensive}\\
      \bottomrule[1.5pt]
    \end{tabular}
    
\end{table*}

\subsubsection{Molecular Docking Experiment}
\begin{figure}[htbp]
    \centering
    \includegraphics[width=0.95\linewidth]{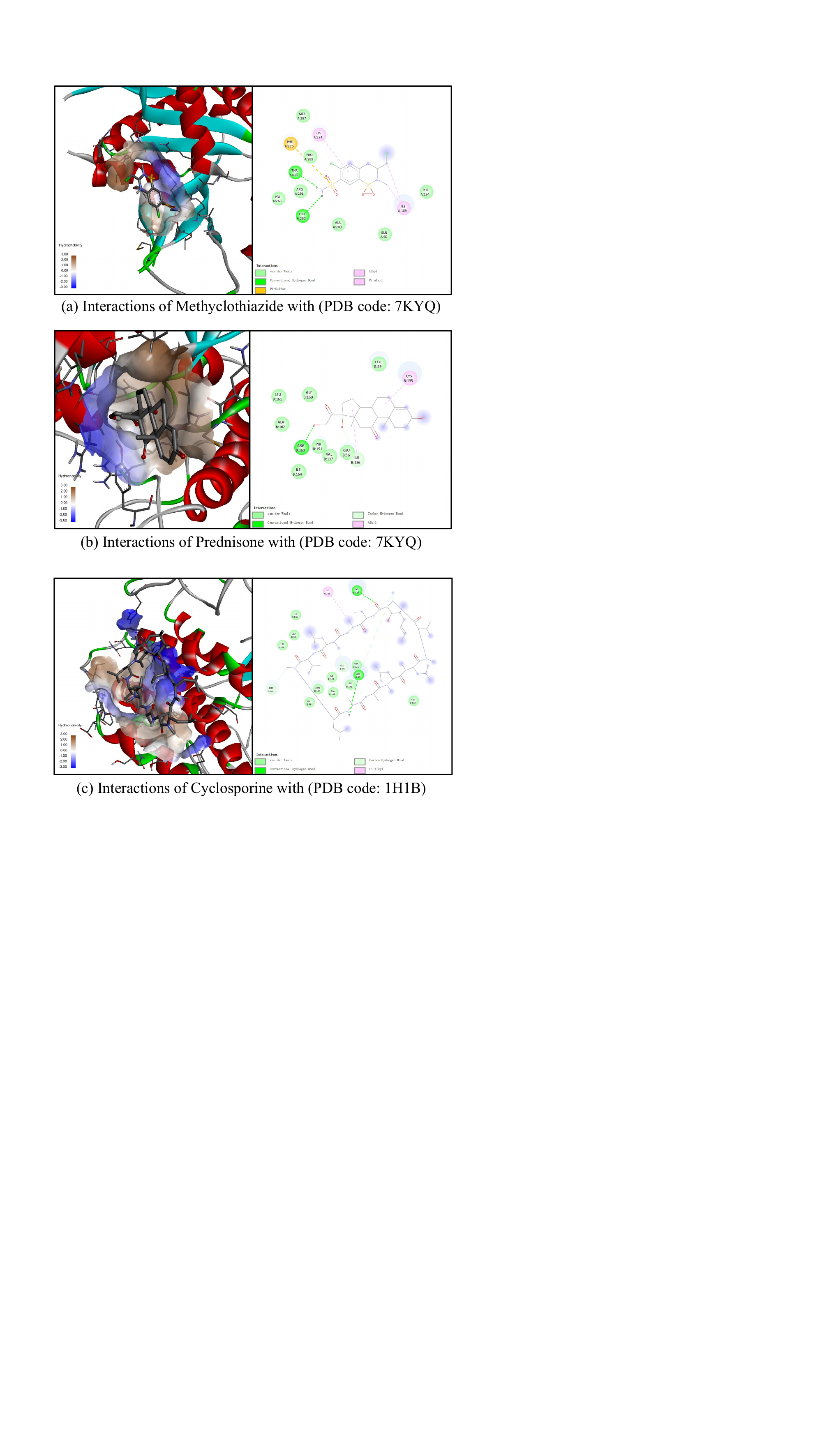}
    \caption{The molecular docking results of ligand molecules with target proteins.}
    \label{Interaction}
\end{figure}

For the unconfirmed drug-disease associations listed in Tab.~\ref{CS}, molecular docking experiments are performed using Autodock Vina and Discovery Studio software to investigate the interaction between drug ligands and target proteins. The binding free energies were calculated to evaluated the pharmacological relevance. The specific molecular docking experiment is illustrated in Fig.~\ref{Interaction}. In the prediction of association between Methyclothiazide and lung cancer, BCCIP (PDB Code: 7KYQ) is selected as the target protein. Fig.~\ref{Interaction}(a) illustrates the molecular interactions between the hydrochlorothiazide ligand and specific amino acid residues of BCCIP (MET287, PRO289, ARG291, VAL268, ALA249, GLN80, and PHE184), including van der Waals forces, conventional hydrogen bonds between nitrogen atoms and residues (THR255, LEU290), and a $\pi-$sulfur interaction (PHE226). The ligand-protein complex exhibited a binding free energy of -11.25 kcal/mol. In the prediction of association between Prednisone and lung cancer, CDK4 (PDB Code: 2W9Z) is selected as the target protein. Fig.~\ref{Interaction}(b) illustrates the van der Waals interactions between the prednisone ligand and specific amino acid residues of CDK4 (GLY160, LEU161, ALA162, ILE164, TYR191, VAL137, GLU56, and LEU59), a conventional hydrogen bond formed between an oxygen atom and residues (ARG163), and a carbon-hydrogen bond (ILE136). The ligand-protein complex exhibited a binding free energy of -14.9397 kcal/mol. In the prediction of association between Cyclosporin and hypertension, the Hexameric (PDB Code: 1P9M) is selected as the target protein. Fig.~\ref{Interaction}(c) illustrates the van der Waals interactions between the cyclosporine ligand and specific amino acid residues of the Hexameric (ILE136, LEU92, THR138, THR149, ILE123, ASN144, ALA145, VAL96, and ASN103), conventional hydrogen bonds formed between oxygen and nitrogen atoms and the residues (LYS120 and GLU99), and a carbon-hydrogen bond (GLU95). Although some of the potentially effective drug-disease associations predicted by BibLDR have not yet been validated by existing evidence, molecular docking experiments and the calculated binding free energies demonstrate the potential value of these predicted associations by BiBLDR.

\section{Conclusion}\label{Section4}
In this paper, we propose a bidirectional behavior learning strategy for drug repositioning. Unlike previous graph representation-based deep learning methods, we frame the drug repositioning task as a recommendation system problem, leveraging behavioral sequence analysis to predict drug-disease associations. We present a novel bidirectional behavioral sequence construction method that simultaneously considers the behavioral sequences of both drug and disease entities, enriching the semantic representation of these sequences. Additionally, we incorporate the Transformer architecture to further model the interactions between different behavioral sequences. Extensive experiments demonstrate that our proposed BiBLDR achieves state-of-the-art performance. cold-start experiments highlight BiBLDR's capability to identify potential indications for new drugs, while tests in sparse environments show that BiBLDR maintains satisfactory performance even with extremely limited data samples. Case studies further illustrate BiBLDR's ability to discover therapeutic candidate drugs for diseases with unknown drug-disease associations. 

In the future, we intend to capture global similarity during prototype construction to shape entity prototypes, rather than focusing solely on pairwise entity relationships. The representation of prototypes in the feature space also warrants further exploration. Although cosine similarity reflects directional differences between vectors, incorporating. Although cosine similarity reflects directional differences between vectors, incorporating richer semantic information may enhance the representational capacity of the prototypes.  In the future processing of bidirectional behavioral sequences, it is necessary to further explore explicit fusion strategies for similarity information as well as the integration of drug-disease associations scores. Moreover, more comprehensive experiments can be conducted to examine how different attention mechanisms influences the processing of bidirectional behavioral sequences.

\section*{References}

\bibliographystyle{IEEEtran}
\bibliography{ref}

\end{document}